\documentclass{optica-article}

\journal{opticajournal} 

\articletype{Research Article}

\newcommand{\image}{{\mathrm{\textbf{I}}}}

\newcommand{\npixels}{\mathbf{N}}

\newcommand{\illum}{L}
\newcommand{\illumone}{L_u}
\newcommand{\illumtwo}{L_v}

\newcommand{\imageone}{\image_{\illumone}}
\newcommand{\imagetwo}{\image_{\illumtwo}}
\newcommand{\Simageone}{\image_A}
\newcommand{\Simagetwo}{\image_B}

\newcommand{\mappinguv}{\mathbf{T}_{\illumone \rightarrow \illumtwo}}
\newcommand{\Smapping}{\mathbf{T}_{A \rightarrow B}}
\newcommand{\mappinguvPredicted}{\hat{\mathbf{T}}_{\illumone \rightarrow \illumtwo}}
\newcommand{\SmappingPredicted}{\hat{\mathbf{T}}_{A \rightarrow B}}

\newcommand{\R}{\mathbb{R}}

\begin{document}

\title{Improved Mapping Between Illuminations and Sensors for RAW Images}

\author{Abhijith Punnappurath,\authormark{1,$\dag$,*}
Luxi Zhao,\authormark{1,$\dag$} Hoang Le,\authormark{2,$\ddagger$} Abdelrahman Abdelhamed,\authormark{3,$\mathsection$} SaiKiran Kumar Tedla,\authormark{2,$\ddagger$} and Michael S. Brown\authormark{1}}

\address{\authormark{1}AI Center-Toronto, Samsung Electronics, 101 College St, Suite 420, Toronto, Ontario M5G 1L7, Canada.\\
\authormark{2}Department of Electrical Engineering and Computer Science, York University, Toronto, Ontario, Canada.\\
\authormark{3}Google Research, 111 Richmond St West, Toronto, ON M5H 2G4, Canada.\\
\authormark{$\dag$}The authors contributed equally to this work.\\
\authormark{$\ddagger$}Work done while an intern at the AI Center-Toronto, Samsung Electronics.\\
\authormark{$\mathsection$}Work done while with the AI Center-Toronto, Samsung Electronics.\\
\email{\authormark{*}abhijith.p@samsung.com}} 

\begin{abstract*} 
RAW images are unprocessed camera sensor output with sensor-specific RGB values based on the sensor's color filter spectral sensitivities. RAW images also incur strong color casts due to the sensor's response to the spectral properties of scene illumination.   The sensor- and illumination-specific nature of RAW images makes it challenging to capture RAW datasets for deep learning methods, as scenes need to be captured for each sensor and under a wide range of illumination. Methods for illumination augmentation for a given sensor and the ability to map RAW images between sensors are important for reducing the burden of data capture. To explore this problem, we introduce the first-of-its-kind dataset comprising carefully captured scenes under a wide range of illumination. Specifically, we use a customized lightbox with tunable illumination spectra to capture several scenes with different cameras. Our illumination and sensor mapping dataset has 390 illuminations, four cameras, and 18 scenes. Using this dataset, we introduce a lightweight neural network approach for illumination and sensor mapping that outperforms competing methods. We demonstrate the utility of our approach on the downstream task of training a neural ISP. Link to project page: \!\href{https://github.com/SamsungLabs/illum-sensor-mapping}{https://github.com/SamsungLabs/illum-sensor-mapping}.

\end{abstract*}

\section{Introduction}
\label{sec:intro}
Preparing image datasets for training and testing is critical to deep-learning methods in computer vision. This task is particularly troublesome for methods targeting RAW images. Most computer vision tasks operate on processed images (i.e., camera-rendered images) that are white-balanced and processed to a device-independent color space (e.g., sRGB). Unlike processed sRGB images, RAW images are unrendered images directly captured by the sensor. RAW images are in a sensor-specific color space based on the spectral sensitivities of the color filter array. Moreover, RAW images are not white-balanced and exhibit a strong response to scene illumination.  Hence, RAW datasets typically need to be captured per sensor and under a wide range of illuminations.

\begin{figure}
  \centering
  \includegraphics[width=0.9\linewidth]{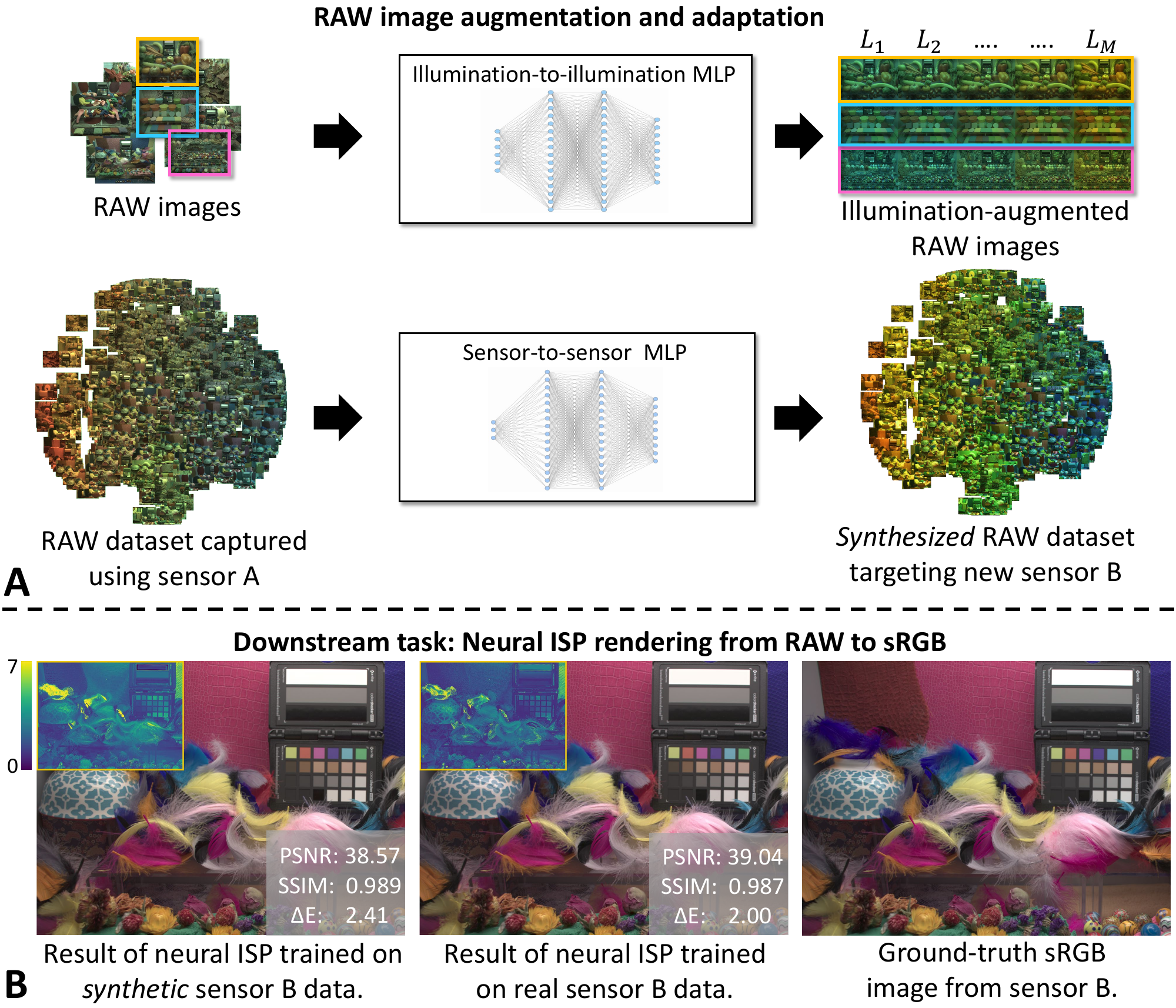}
  \caption{(A) Our illumination mapping approach can convert a RAW image captured under a source illumination to appear as captured under an arbitrary target illumination. Our sensor mapping method can convert a RAW image captured by a source sensor to a target sensor. (B) An example showing the utility of our illumination and sensor mapping approaches for use in neural ISP training. A neural ISP trained on our synthetic dataset offers competitive performance compared to training on real images.  Inset shows $\Delta$E~\cite{deltaE} error map.
  }\label{fig:Teaser}
\end{figure}

To reduce the burden of RAW image capture, data augmentation methods tailored for RAW images are critical.  In this paper, we examine two methods: (1) illumination-to-illumination mapping that converts a RAW image from a particular sensor captured under a source illumination to a target illumination; (2) sensor-to-sensor mapping that converts a RAW image captured by a source sensor under some illumination to appear as if a target sensor captured it under the same illumination. See Fig.~\ref{fig:Teaser}-A. 

Several ad-hoc augmentation approaches have been proposed for various low-level computer vision tasks that work directly with RAW images. Often, a color checker rendition chart (color chart) is used to provide RAW color samples from different illumination and sensors. Prior works often exploit the availability of the color chart, which is often placed in the scene for purposes other than data augmentation.    This paper aims to thoroughly study the illumination-to-illumination and sensor-to-sensor problems without relying on existing datasets. To this end, we have prepared a new dataset for this purpose. Our dataset consists of 18 scenes (13 real-object scenes, 4 scenes with custom color charts, and 1 scene with a standard X-Rite color chart) captured by four cameras using 390 different illuminations. RAW images were carefully captured in a GTI lightbox with a tunable light source comprised of narrow-band LEDs that replicate the spectral power distributions of commercial and natural lighting. In total, our dataset has over 8500 images for testing and training.

Using this new dataset, we examine the illumination-to-illumination and sensor-to-sensor mapping problems. One of our observations is that relying on a single X-Rite color chart does not provide suitable material samples to address these problems.  Instead, we advocate using more samples derived directly from materials indicative of real scenes.  Using dense samples, we propose a multilayer perceptron (MLP) method to predict the necessary transforms for illumination-to-illumination and sensor-to-sensor mapping. We show that our MLP with just 1481 parameters and a model size of only 32 KB outperforms alternative strategies.

We demonstrate the utility of our illumination and sensor mapping approaches for the downstream task of neural rendering. 
Cameras render RAW sensor images to sRGB through a series of operations performed by hardware called an image signal processor (ISP).  A recent trend is to replace conventional ISPs with neural ISPs. Training a neural ISP requires a dataset of RAW sensor-specific images captured under a wide variety of illuminations.   Many smartphones now have multiple cameras per device, requiring data capture for each sensor. Given an existing dataset captured for a particular sensor, our illumination and sensor mapping methods can be used to generate synthetic training data for a neural ISP targeting a different sensor. As shown in Fig.~\ref{fig:Teaser}-B, a neural ISP trained on our synthetic data offers competitive performance to training on real images from the target sensor, representing significant savings in time and effort.

\section{Related work}
\label{sec:relatedworks}

\noindent{\textbf{Conventional color augmentation for processed images.}} Most learning-based methods are trained on images processed to a common color space---namely standard RGB (sRGB). Color augmentation strategies are used to produce variations in the training data. One well-known example is RGB color jittering, which involves independently scaling the RGB color channels with scaling factors represented by a $3\!\times\!3$ diagonal matrix. Other common color augmentation methods include PCA-based shifting, HSV jittering, color channel dropping and swapping. Such methods are widely used in machine-learning methods and libraries ~\cite{chatfield2014return, cubuk2018autoaugment, doersch2015unsupervised, kalantari2017deep, krizhevsky2017imagenet, lee2017unsupervised, movshovitz2016useful, redmon2016you, jung2019imgaug}. While these methods may be suitable for sRGB images, given sRGB images are encoded in a standardized color space, they are unrealistic for RAW.
RAW images are the response of the imaging sensor and have a direct relationship to the scene reflectance, scene illumination, and the sensor's spectral sensitivities. The sensor-specific nature of RAW images means different sensors' RGB responses to the same scene can be vastly different.

\noindent{\textbf{RAW illumination augmentation with illuminant vectors.}}~Interestingly, augmentation methods for RAW images have not been significantly different than those applied to processed images. For example, FC4~\cite{hu2017fc} randomly samples an illuminant vector measured from an existing dataset and changes the illuminant of an image by multiplying the image by a \textit{diagonal $3\!\times\!3$ matrix} containing this illuminant vector. This strategy guarantees correct values only on achromatic scene materials (i.e., grey/white objects) and does not guarantee correct color values for non-neutral materials. As a result, the augmented image does not depict a realistic color distribution that would have been obtained by changing the scene illumination.

\noindent{\textbf{RAW illumination and sensor mapping with color charts.}}~More accurate illumination mapping methods (e.g.,~\cite{lo2021clcc,abdelhamed2021leveraging}) use X-Rite color rendition charts that contain 24 color patches representing a distinct distribution of colors. Such color charts are commonly used in color constancy and illuminant estimation datasets. The CLCC method~\cite{lo2021clcc} performs illumination augmentation by estimating a $3\!\times\!3$ transformation matrix between the color charts found under two illuminants in different scenes. Despite being more accurate than using a diagonal matrix, the estimated $3\!\times\!3$ matrix is biased towards correcting materials that are similar to the X-Rite color checker samples. As we will show, these color patches do not represent most real materials well. Similarly, work by~\cite{nguyen2015raw} proposed a sensor-to-sensor mapping approach that relied solely on color charts to provide color samples among different sensors. Again, like the illumination-to-illumination task, we observe that X-Rite color chart samples are insufficient for the sensor-to-sensor mapping tasks.

One of the drawbacks of previous illumination and sensor mapping methods for RAW images is the need for a comprehensive image dataset to address this problem properly. In the following, we describe our RAW dataset capture and our proposed illumination and sensor mapping method.

\section{Dataset capture}\label{sec:dataset}

A RAW image dataset of scenes with colorful materials captured under a dense illumination sampling by multiple cameras is needed to study the illumination and sensor mapping problem.  Our dataset intends to provide a rich representation of different camera sensor responses to a wide variety of materials and illuminations.  In addition, we wanted careful image capture to ensure the entire scene was lit by a single uniform illumination.  The following section details our dataset collection procedure.

\begin{figure}[!t]
  \centering
  \includegraphics[width=0.9\linewidth]{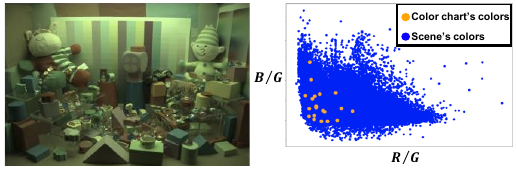}
  \caption{Left: Our training scene. Right: Distribution of colors for the Canon EOS Rebel SL2 camera under the D65 illuminant from a color chart versus our training scene on the left, plotted in the $[R/G,B/G]$ chromaticity space.
    \label{fig:color-chart-vs-scene}
  }
\end{figure}

\noindent{\textbf{Lightbox and tunable lightsource.}}~Images were captured using a GTI lightbox~\cite{gti} that had its original light sources removed and replaced with a Telelumen's multispectral luminaire light source~\cite{telelumens}. The Telelumen's luminaire and its associated software are comprised of narrow-band LEDs that can be adjusted to match a specified broad-spectrum power distribution (SPD) to emulate virtually any light source. This allows us to specify a target SPD for the lightbox. Using the tunable lightbox to control light sources' SPD, we capture the same scene under many different illuminants.  

\begin{figure}[t]
  \centering
  \includegraphics[width=0.8\linewidth]{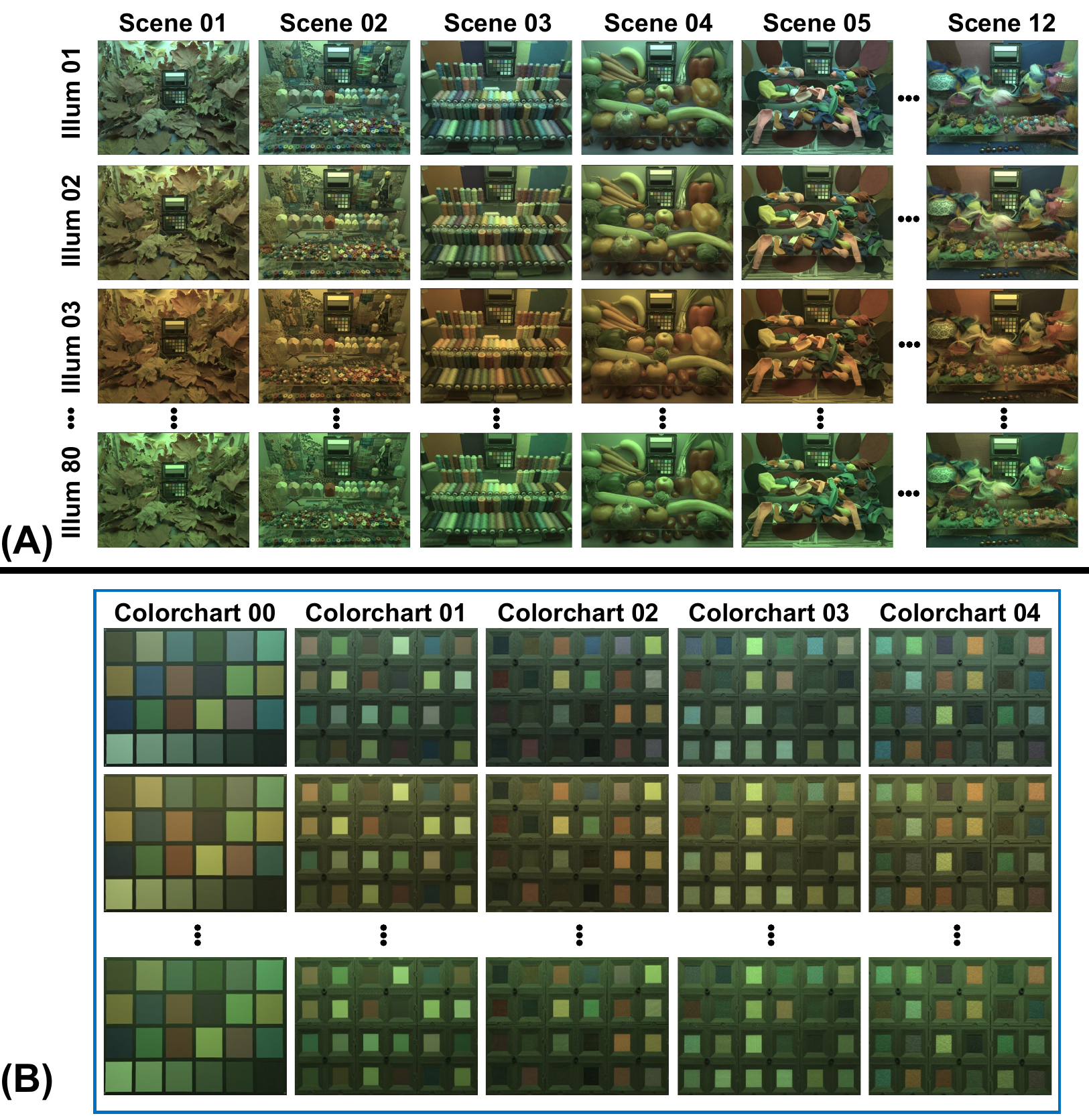}
  \caption{(A) Illumination-to-illumination mapping dataset. Representative examples from our 12 test scenes captured under 80 test illuminants are shown. (B) Sensor-to-sensor mapping dataset. Examples of our customized color charts with real materials. {\it Colorchart 00} shows the standard X-Rite chart.  Both (A) and (B) show examples from the Samsung S22 Plus smartphone.
    \label{fig:dataset}
  }
\end{figure}

We collected spectral data from multiple sources~\cite{SFU,PhotoLED,LSPDD,RealLightSource}, which consists of more than 3,000 SPDs, plus the simulated canonical light D65 from the lightbox. From this assimilated SPD database, we randomly select 390 different illuminants that include most real-world light sources (e.g., sunlight, LED, compact fluorescent, and incandescent). We split the 390 illuminants into 250 illuminants for training, 60 for validating, and 80 for testing. Note that we include the simulated D65 SPD into the \textit{train} split for other baseline methods. We normalize the power of all SPDs to 800 Watts to ensure no SPD exceeds the lightbox's capability.

With the lightbox and a dataset of illuminants' SPDs, we sweep through all illuminants and use four different cameras (3 smartphones and 1 DSLR) to capture scenes. Our four cameras are Samsung S22 Plus, Samsung S22 Ultra, Google Pixel 6, and a Canon EOS Rebel SL2. We use the lowest ISO on the camera to minimize the effect of noise. For our training and validation scene, we only built one scene with a fixed set of objects (colorful papers, plastic and wooden toys) which is captured with our four cameras under 250 training illuminants and 60 validating illuminants. The left column of Fig.~\ref{fig:color-chart-vs-scene} shows our training scene and the chromaticity plot on the right shows the distribution of colors from a color chart versus our training scene under the D65 illuminant for the Canon EOS Rebel SL2 camera. As can be observed, the X-Rite color chart values span only a limited portion of the chromaticities of real materials. We also capture a scene with only a color chart to extract all 390 illuminants in RAW-RGB values per camera.

We also designed another 12 natural scenes and captured them under 80 test illuminants. Fig.~\ref{fig:dataset}-A shows examples from our captured test scenes for the Samsung S22 Plus. Note that the other three cameras also capture the same test scenes. The test scenes exclude the objects used in the training scene. In our test scenes, we choose objects made of various materials, including artificial and natural materials. We conduct our experiments inside a closed darkroom.
The scenes are static, and the camera is mounted on a tripod. RAW-DNG files are saved using the Android Camera2 API for the three smartphones. The Canon DSLR's RAW CR2 files are converted to DNG using the Adobe DNG Converter.

This dataset allows quantitative evaluation of our illumination-to-illumination mapping method because we have aligned images of the same scene under different illuminations. However, it is difficult to register images across different cameras because the fields of view are different, and the camera centers cannot be precisely aligned. While additional hardware, such as beamsplitters, may alleviate this issue, such setups introduce their own challenges in terms of design complexity and calibration. Therefore, we use customized charts built from real materials for the sensor-to-sensor mapping experiment. Fig.~\ref{fig:dataset}-B provides a few representative examples. We create four color charts using 3D-printed frames to ensure the materials are flat. We choose various materials---paint chips, paper, suede, and cotton cloth---that are homogeneous in terms of color. In total, we have five color charts, including the X-Rite chart. The four custom charts have 24 colors, similar to the X-Rite chart. There are no shared materials among the custom charts. We manually cropped the 24 colored patches and averaged each patch to produce 24 color samples. Our experiments use only the individual chart's colors to evaluate our sensor-to-sensor mapping algorithm.
Similar to the illumination mapping dataset, we image a single training scene, where the scene comprises a custom color chart. This custom chart scene is also captured under 250 training illuminants. We capture another custom color chart comprised of different materials under 60 validating illuminants and the remaining two custom charts under 80 testing illuminants.

\begin{figure*}[t]
  \centering
  \includegraphics[width=\linewidth]{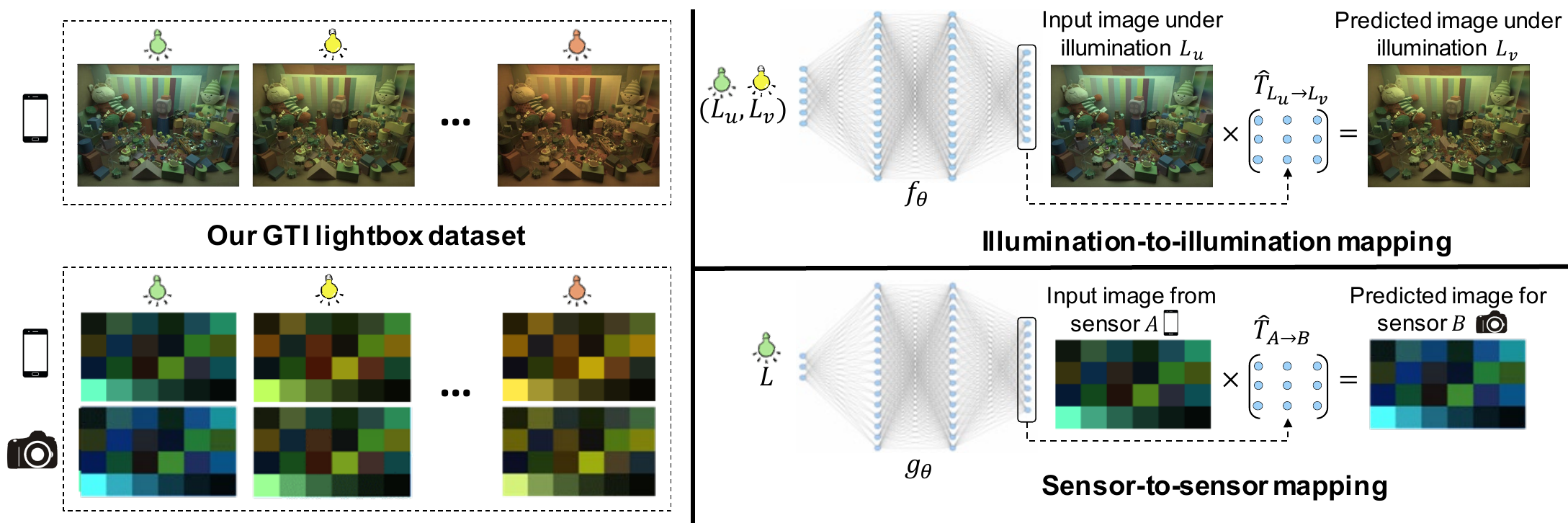}
  \caption{Overview of our illumination-to-illumination and sensor-to-sensor mapping approaches. For illumination mapping, we train an MLP that takes the source and target illuminations as input and predicts the $3\!\times\!3$ transformation matrix that maps the input image under the source illumination to the target illumination. A similar approach is applied for sensor mapping where the MLP, conditioned on the illumination, predicts the transformation that maps the input image from the source sensor to the target sensor's color space. Our GTI lightbox dataset comprising images from multiple sensors spanning diverse scenes captured under different controlled illuminations provides the necessary data for training.
    \label{fig:method-overview}
  }
\end{figure*}

\section{Illumination-to-illumination mapping}
\label{illuminant-mapping}
Given a demosaiced RAW-RGB image, $\imageone \in \R ^ {3 \times \npixels}$, captured under illuminant $\illumone \in \R ^ 3$, we want to transform the image such that it appears as if the same scene was captured under a different illuminant $\illumtwo$ using the same sensor (the target image is denoted as $\imagetwo$), where $\npixels$ is the number of RGB pixels in the image and the illuminants are represented as [R, G, B] triplets. This transformation from $\imageone$ to $\imagetwo$ can be expressed as follows:
\begin{equation}\label{eq:illuminant-transform}
  \imagetwo = \mappinguv \imageone,
\end{equation}
where $\mappinguv \in \R ^ {3\!\times\!3}$ is a transformation matrix that maps the RAW-RGB image $\imageone$ to $\imagetwo$.

For a specific camera sensor, we use a multi-layer perceptron (MLP) that takes as input two illuminants, a source illuminant $\illumone$ and a target illuminant $\illumtwo$, then estimates a $3\!\times\!3$ transformation matrix $\mappinguvPredicted$, as:
\begin{equation}
  \mappinguvPredicted = f_{\theta}(\illumone, \illumtwo),
\end{equation}
where $f_{\theta}$ refers to the MLP model with the parameters $\theta$. Our MLP consists of two hidden layers, each consisting of 32 units. The MLP outputs nine parameters reshaped to obtain the $3\!\times\!3$ transformation matrix, as shown in Fig.~\ref{fig:method-overview}.

Note that Eq.~\ref{eq:illuminant-transform} is different from conventional white balancing~\cite{cheng2015}.
White balancing is typically performed using a diagonal $3\!\times\!3$ matrix under the assumption that the three RGB channels of the camera sensor act as independent gain controls to scene illumination, and therefore, a diagonal matrix can be used to correct the three RGB channels by normalizing their individual channel bias. While the diagonal model is effective at correcting neutral colors (e.g., achromatic greys/whites), there can be errors in non-neutral colors~\cite{cheng2015}. In contrast, our method uses a full $3\!\times\!3$ matrix to correct {\it all} colors, and not only the neutral colors. Also, $\illumone$ and $\illumtwo$ can be any arbitrary illuminants in the RAW-RGB color space of the imaging sensor.

To train our illumination mapping MLP, we use the \textit{training scene} from our dataset, containing various colorful materials, and captured under 310 different illuminants (250 for training and 60 for validation). The various colorful materials and illuminants provide denser representations of the illuminant space of the underlying imaging sensor. To ensure pixel-wise correspondence between image pairs due to small vibrations during imaging, we downscale the images by a factor of 4 for the three smartphone sensors and by a factor of 6 for the Canon DSLR, using bilinear interpolation. Let us denote the downsampled images representing this scene as $\{ \image_{\illum_t} \}$, where $t \in \{ 1, \dots, K \}$ and $K$ is the number of training illuminants. $K = 250$ for our training scene. From each image pair $\{ \image_{\illumone}, \image_{\illumtwo} \}$, we randomly obtain a subset of 1000 corresponding pixels $\{ \image^{\prime}_{\illumone},\image^{\prime}_{\illumtwo} \}$.

To train our MLP, we use angular error between target image samples $\image^{\prime}_{\illumtwo}$ and estimated image samples $\hat{\image}^{\prime}_{\illumtwo}$ as the loss function, where $\hat{\image}^{\prime}_{\illumtwo}$ is obtained by applying the MLP's prediction $\mappinguvPredicted$ on source image samples $\image^{\prime}_{\illumone}$ as   $\hat{\image}^{\prime}_{\illumtwo} = \mappinguvPredicted \image^{\prime}_{\illumone}$. At inference time, the transformation matrix $\mappinguvPredicted$ is applied to the full image $\image_{\illumone}$ to obtain $\hat{\image}_{\illumtwo}$.

\section{Sensor-to-sensor mapping}
\label{sensor-mapping}

Analogous to the illumination mapping approach, where we map from a source illuminant to a target illuminant for a given sensor, we use a similar approach to map from a source sensor to a target sensor under a given illuminant. Formally, given a demosaiced RAW-RGB image, $\Simageone \in \R ^ {3 \times \npixels}$, captured using sensor $A$ under illuminant $\illum \in \R ^ 3$, we want to transform the image such that it appears as if the same scene was captured using sensor $B$ under the same lighting condition. The transformation from $\Simageone$ to $\Simagetwo$, where $\Simagetwo$ represents the target image from sensor $B$, can be expressed as:
\begin{equation}\label{eq:sensor-transform}
  \Simagetwo = \Smapping \Simageone,
\end{equation}
where $\Smapping \in \R ^ {3\!\times\!3}$ is a transformation matrix that maps the RAW-RGB image $\Simageone$ to $\Simagetwo$. Note that the transformation $\Smapping$ varies with the illumination $\illum$.

We train a separate MLP for each source-target sensor pair. The MLP takes the illuminant $\illum$ as input and estimates a $3\!\times\!3$ transformation matrix $\SmappingPredicted$ as:
\begin{equation}
  \SmappingPredicted = g_{\theta}(\illum),
\end{equation}
where $g_{\theta}$ refers to the MLP model with the parameters $\theta$. We use the same MLP architecture as before, with two hidden layers and 32 units per layer.

We use angular error between the target image samples $\image^{\prime}_{B}$ and the estimated image samples $\hat{\image}^{\prime}_{B}$ to train our MLP, where $\hat{\image}^{\prime}_{B}$ is obtained by applying the MLP's prediction $\SmappingPredicted$ on the source image samples $\image^{\prime}_{A}$ as   $\hat{\image}^{\prime}_{B} = \SmappingPredicted \image^{\prime}_{A}$. The samples $\{ \image^{\prime}_{A}, \image^{\prime}_{B} \}$ are the 24 values from the custom color charts. Note that our focus is on accurately transforming the sensor-specific RAW colors from the source domain to the target domain. As as a result, modeling of other sensor-specific effects, notably noise, is not addressed here. Noise can be modeled using sensor-specific noise models.

\section{Experiments}

\subsection{Implementation details}
For illuminant-to-illuminant mapping experiments, we train our MLP using the Adam optimizer~\cite{adam} with $\beta_1=0.9$ and $\beta_2=0.999$. The learning rate starts at 0.01 and is decayed by 0.5 every 50 epochs. All models are trained using PyTorch for 400 epochs with a batch size of 8. Since we use only an angular error loss, which computes the angle between the two vectors, and is independent of their magnitude, we normalize the transformation matrix output by the MLP both during training and testing. To learn a robust model, we train only on `hard' examples by selecting source/target illumination pairs that are far away in $\left[\frac{R}{G}, \frac{B}{G}\right]$ chromaticity space. Based on this criterion, we select 17566 illuminant pairs for training, 976 illuminant pairs for validating, and 2010 illuminant pairs for testing. We apply a few basic preprocessing operations to the RAW-DNG files to obtain the training data $\{ \image_{\illum_t} \}$. In particular, we apply a black-/white-level correction, demosaicing, downsampling, and normalization. The subset of 1000 pixels $\{ \image^{\prime}_{\illum_t} \}$ is randomly sampled at each epoch of training. For sensor-to-sensor mapping experiments, we train our MLP using the same optimizer but with a learning rate of 0.001.    

\subsection{Illumination-to-illumination mapping}
\label{sec:illum_mapping_expts}
We compare our method to the following baselines: (1) diagonal matrix; (2) KNN-based full matrix; (3) image translation using a U-Net~\cite{unet}; and (4) an oracle selecting the optimal transform.

\noindent{\textbf{Diagonal matrix.}} We apply a diagonal correction where we divide the source image by the source illumination and then multiply it by the target illumination. This can be implemented as matrix multiplication using two diagonal matrices. Diagonal correction can correct neutral colors but may produce erroneous results on non-neutral colors.

\begin{figure*}[t]
  \centering
  \includegraphics[width=1\linewidth]{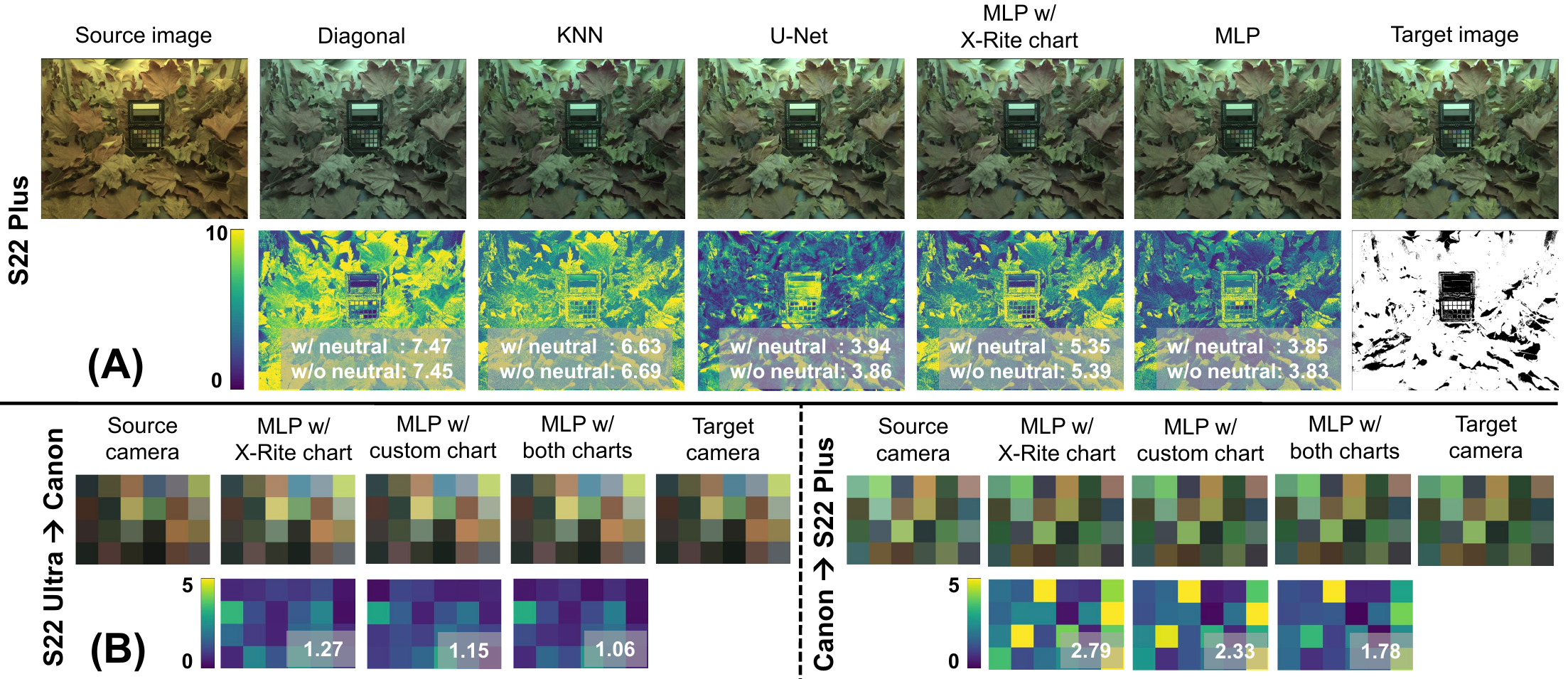}
  \caption{(A) Qualitative comparisons between our method and competing approaches for illumination-to-illumination mapping. The binary mask shows the non-saturated non-neutral pixels over which the angular error is computed in white color. The third KNN variant 2NN--1NN~\cite{lo2021clcc} is shown. The error map shows per-pixel angular error, and insets provide the mean angular error. (B) Qualitative comparisons between our MLP variants for sensor-to-sensor mapping.
    \label{fig:results-visual}
  }
\end{figure*}

\noindent{\textbf{KNN-based mapping.}} We also compare against a K-nearest neighbor (KNN) method. We assume that the transformations $\mappinguv$ for all $\{ \illumone, \illumtwo \}$ pairs in the training set are pre-computed and available at test time. Given a source and/or target illuminant not belonging to the training set, we compute the Euclidean distance in $\left[ \frac{R}{G}, \frac{B}{G} \right]$ chromaticity space from the test illuminant to its K nearest neighbors in the training set $\{ \illum_t \}$. To obtain the transformation matrix, we compute a weighted average of the corresponding nearest neighbor transformations $\mappinguv$, with the normalized inverse Euclidean distances applied as the weights. Depending on the value of K used for interpolation at source and target, we consider four KNN variations---(i) 1NN--1NN, (ii) 1NN--KNN, (iii) KNN--1NN, and (iv) KNN--D65--KNN. The first variation is simply selecting the transformation $\mappinguv$ where $\illumone$ and $\illumtwo$ are the closest illuminants in the training set to the test time source and target illuminants. We use K$=2$ for the remaining variations. Note that (i) 1NN--1NN is similar to the approach adopted by~\cite{nguyen2015raw} for sensor mapping. CLCC~\cite{lo2021clcc} used the strategy of (iii) KNN--1NN for illumination augmentation. 

For (iv) KNN--D65--KNN, we follow a two-step approach where we estimate a first transform using KNN for the source illumination that maps to daylight D65 illumination. The training set includes the D65 illumination. Next, we compute a second transform that maps D65 to the target illumination, with KNN applied to the target illumination.
  
\noindent{\textbf{U-Net mapping.}} We also compare against a deep learning solution using a standard U-Net architecture~\cite{unet}. The input to the U-Net contains seven channels, with the first three channels being the RAW source image, and the fourth and the fifth, and the sixth and the seventh channels are the $\frac{R}{G}$ and $\frac{B}{G}$ chromaticity values of the source and target illumination, respectively, tiled to the image width and height. The U-Net is trained to predict the three-channel RAW target image. Please refer to the supplementary materials for more implementation details. 

\noindent{\textbf{Oracle.}} Finally, we also compare against an {\it oracle} method. This result is obtained by leveraging our MLP framework. Specifically, we finetune our trained MLP on \textit{each given testing pair} for 200 epochs with a learning rate of 0.001. This represents the best $3\!\times\!3$ transform matrix, in terms of angular error, that maps the image under the source illumination to the target illumination.  The intention of this oracle method is to demonstrate the best a  $3\!\times\!3$ matrix can achieve for illumination-to-illumination mapping.

\begin{table*}[t]
  \caption{Quantitative results for illumination-to-illumination mapping. Mean angular errors (MAE) are reported. We report results over all pixels `w/ ntrl' and excluding the neutral pixels `w/o ntrl'.}
  \centering
  \label{tab:illuminant-mapping-results}
  \resizebox{0.8\textwidth}{!}{%
  \setlength{\tabcolsep}{6pt}
  \begin{tabular}{l|cc|cc|cc|cc|}
  \cline{2-9}
   &
    \multicolumn{2}{c|}{\textbf{S22 Plus}} &
    \multicolumn{2}{c|}{\textbf{S22 Ultra}} &
    \multicolumn{2}{c|}{\textbf{Pixel 6}} &
    \multicolumn{2}{c|}{\textbf{Canon}} \\ \hline
  \multicolumn{1}{|l|}{\textbf{Method}} &
    \multicolumn{1}{c|}{\textbf{w/ ntrl}} &
    \textbf{w/o ntrl} &
    \multicolumn{1}{c|}{\textbf{w/ ntrl}} &
    \textbf{w/o ntrl} &
    \multicolumn{1}{c|}{\textbf{w/ ntrl}} &
    \textbf{w/o ntrl} &
    \multicolumn{1}{c|}{\textbf{w/ ntrl}} &
    \textbf{w/o ntrl} \\ \hline
  \multicolumn{1}{|l|}{Diagonal} &
    \multicolumn{1}{c|}{4.51} &
    5.01 &
    \multicolumn{1}{c|}{4.87} &
    5.39 &
    \multicolumn{1}{c|}{5.12} &
    5.76 &
    \multicolumn{1}{c|}{3.89} &
    4.53 \\ \hline
  \multicolumn{1}{|l|}{KNN 1NN--1NN} &
    \multicolumn{1}{c|}{4.89} &
    4.90 &
    \multicolumn{1}{c|}{6.14} &
    6.25 &
    \multicolumn{1}{c|}{6.42} &
    6.72 &
    \multicolumn{1}{c|}{4.51} &
    4.54 \\ \hline
  \multicolumn{1}{|l|}{KNN 1NN--2NN} &
    \multicolumn{1}{c|}{4.93} &
    4.93 &
    \multicolumn{1}{c|}{6.16} &
    6.26 &
    \multicolumn{1}{c|}{6.47} &
    6.76 &
    \multicolumn{1}{c|}{4.50} &
    4.53 \\ \hline
  \multicolumn{1}{|l|}{KNN 2NN--1NN} &
    \multicolumn{1}{c|}{4.87} &
    4.87 &
    \multicolumn{1}{c|}{5.89} &
    6.00 &
    \multicolumn{1}{c|}{6.37} &
    6.67 &
    \multicolumn{1}{c|}{4.51} &
    4.54 \\ \hline
  \multicolumn{1}{|l|}{KNN 2NN--D65--2NN} &
    \multicolumn{1}{c|}{5.20} &
    5.20 &
    \multicolumn{1}{c|}{6.86} &
    6.84 &
    \multicolumn{1}{c|}{6.93} &
    7.35 &
    \multicolumn{1}{c|}{4.76} &
    4.67 \\ \hline
  \multicolumn{1}{|l|}{U-Net} &
    \multicolumn{1}{c|}{5.41} &
    5.44 &
    \multicolumn{1}{c|}{6.20} &
    6.20 &
    \multicolumn{1}{c|}{5.04} &
    5.34 &
    \multicolumn{1}{c|}{3.79} &
    3.89 \\ \hline
  \multicolumn{1}{|l|}{Our MLP w/ color chart} &
    \multicolumn{1}{c|}{4.61} &
    4.68 &
    \multicolumn{1}{c|}{4.60} &
    4.69 &
    \multicolumn{1}{c|}{6.02} &
    6.31 &
    \multicolumn{1}{c|}{3.58} &
    3.71 \\ \hline
  \multicolumn{1}{|l|}{Our MLP} &
    \multicolumn{1}{c|}{\cellcolor{yellow!35}3.63} &
    \cellcolor{yellow!35}3.67 &
    \multicolumn{1}{c|}{\cellcolor{yellow!35}3.99} &
    \cellcolor{yellow!35}4.08 &
    \multicolumn{1}{c|}{\cellcolor{yellow!35}4.90} &
    \cellcolor{yellow!35}5.18 &
    \multicolumn{1}{c|}{\cellcolor{yellow!35}3.27} &
    \cellcolor{yellow!35}3.33 \\ \hline
  \multicolumn{1}{|l|}{MLP Oracle} &
    \multicolumn{1}{c|}{2.81} &
    2.95 &
    \multicolumn{1}{c|}{3.08} &
    3.26 &
    \multicolumn{1}{c|}{3.82} &
    4.25 &
    \multicolumn{1}{c|}{2.12} &
    2.27 \\ \hline
  \end{tabular}%
  }
  \end{table*}

Quantitative results of illumination mapping are shown in Table~\ref{tab:illuminant-mapping-results}.
We report the mean angular error (MAE) averaged over a total of 2010 illumination pairs covering 12 scenes for each camera. For a fair comparison, we exclude saturated pixels from our MAE computation since angular errors are not meaningful if one or more color channels are saturated. Specifically, we flag pixels as saturated if at least one color channel in the target image is less than 1\% or greater than 99\% of the sensor's dynamic range. In the table, we report results over all pixels labeled as ‘w/ ntrl’ and exclude the neutral pixels (i.e., achromatic materials) denoted as ‘w/o ntrl’. We define neutral pixels as those pixels in the target image whose angular error computed with respect to the target illumination is less than 3.5 degrees. Note that methods such as diagonal correction may work well for neutral colors while being highly erroneous for non-neutral colors. We introduce this split of with and without neutral colors to examine the accuracy of full-color correction in isolation. 

From Table~\ref{tab:illuminant-mapping-results}, it can be observed that the simple diagonal correction approach, as expected, incurs a significantly higher error on non-neutral pixels. For the KNN variants, (iii) 2NN--1NN with interpolation is more accurate than (i) 1NN--1NN, where no interpolation is performed. We also found that the two-step approach in (iv) 2NN--D65--2NN does not perform as well as the other variants because errors in transform estimation may accumulate with the intermediate step at D65. The U-Net is better than the diagonal and KNN on the Pixel 6 and the Canon cameras but performs poorly on the two Samsung devices, with higher errors than the diagonal approach. Our MLP trained on real materials offers the best performance compared to all baselines and has the least gap to the oracle method. Moreover, our MLP is more accurate than an MLP trained only on a color chart, highlighting the need for diverse colors representative of real scene materials.
Representative qualitative results are presented in Fig.~\ref{fig:results-visual}-A.

\subsection{Sensor-to-sensor mapping}
\label{sec:sensor_mapping_expts}

We conduct our sensor mapping experiments with four variants of our MLP: (1) MLP trained on the X-Rite color chart, (2) MLP trained using a custom chart, (3) MLP trained with both an X-Rite chart and a custom chart, and (4) an oracle MLP. For sensor mapping, the number of color samples available for training is limited---only 24 colors per chart as against 1000 samples used for illuminant mapping. Variants (1) and (2) use only 24 colors, while (3) has access to 48 color values. Similar to the previous experiment, the oracle method is obtained by training our MLP on each given testing pair for 200 epochs with a learning rate of 0.001.

\begin{table*}[t]
  \caption{Quantitative results for sensor-to-sensor mapping. Mean angular errors (MAE) are reported.}
      \centering
  \label{tab:sensor-mapping-results}
  \resizebox{0.825\textwidth}{!}{%
   \setlength{\tabcolsep}{6pt}
  \begin{tabular}{|l|c|c|c|c|c|c|}
  \hline
  \textbf{Method} &
    \textbf{S22 Plus to} &
    \textbf{S22 Ultra to} &
    \textbf{S22 Plus to} &
    \textbf{S22 Ultra to} &
    \textbf{Canon to} &
    \textbf{Canon to} \\
       &
    \textbf{S22 Ultra} &
    \textbf{S22 Plus} &
    \textbf{Canon} &
    \textbf{Canon} &
    \textbf{S22 Plus} &
    \textbf{S22 Ultra} \\ \hline

KNN 2NN w/ X-Rite chart & 0.54 & 0.55 & 1.79 & 2.01 & 2.00 & 2.17 \\ \hline
KNN 2NN w/ custom chart  & 0.51 & 0.62 & 1.38 & 1.69 & 1.52 & 1.61 \\ \hline
KNN 2NN w/ both        & 0.51 & 0.55 & 1.58 & 1.82 & 1.71 & 1.80 \\ \hline

  MLP w/ X-rite chart & 0.44 & 0.44 & 1.23 & 1.22 & 1.37 & 1.22 \\ \hline
  MLP w/ custom chart  & 0.41 & 0.43 & 1.21 & 1.21 & 1.32 & 1.12 \\ \hline
  MLP w/ both &
    \cellcolor{yellow!35}0.40 &
    \cellcolor{yellow!35}0.42 &
    \cellcolor{yellow!35}1.10 &
    \cellcolor{yellow!35}1.10 &
    \cellcolor{yellow!35}1.15 &
    \cellcolor{yellow!35}1.09 \\ \hline
  MLP Oracle         & 0.34 & 0.35 & 0.50 & 0.49 & 0.50 & 0.52 \\ \hline
  \end{tabular}%
  }
  \end{table*}

Quantitative results for sensor-to-sensor mapping are presented in Table~\ref{tab:sensor-mapping-results}. A KNN method is also included as a baseline. Qualitative results are presented in Fig.~\ref{fig:results-visual}-B. 
All methods are tested on two custom charts with no materials shared with the training charts. As shown in Table~\ref{tab:sensor-mapping-results}, the MLP trained on our custom chart performs better than the one trained on the X-Rite chart.
The MLP trained on the X-Rite and the custom chart yielded even better performance.

\noindent \textbf{Generalization.} These results 
show that our mapping framework can generalize well to new illuminations and scene content. For a new sensor, our method requires minimal data capture effort. Specifically, to train our MLPs, we need to image a few color charts or colorful scenes under different illuminations. When used as an augmentation strategy for downstream tasks, our method considerably reduces the burden of data capture, as we shall demonstrate next.

\subsection{Neural ISP}
Our final experiment demonstrates the usefulness of our approach for low-level vision problems that work with RAW input images. We consider the scenario
where there is an existing dataset from a particular sensor (A), and we want to train a neural ISP for a new sensor (B) with minimal data capture. For example, we want to collect a dataset for sensor B that is $M$ times smaller than the dataset for sensor A ($M=5$ in our experiments). We use a subset of our lightbox dataset to train and evaluate the neural ISP. We chose this dataset since it already contains many images captured using multiple sensors and allows us to quantitatively evaluate the performance of synthetic models that map from sensor A to sensor B against an `oracle' model that is trained purely on real data from sensor B. In particular, we use the original test split of the illumination-mapping dataset containing 12 scenes under 80 illuminations and divide it as follows---train: 6 scenes / 40 illuminants, validation: 3 scenes / 10 illuminants, test: 3 scenes / 30 illuminants. This ensures that scenes and illuminants are not shared across splits. We use this training and validation split to train the oracle model. 
\begin{table}[t]
\centering
  \caption{Quantitative results on the downstream task of neural rendering. Sensor A source: Samsung S22 Ultra smartphone, sensor B target: Canon DSLR. A lower $\Delta$E~\cite{deltaE} is better.}
      \centering
  \label{tab:neural_isp}
  \resizebox{0.45\columnwidth}{!}{%
  \setlength{\tabcolsep}{12pt}
  \begin{tabular}{|l|c|c|c|}
  \hline
  \textbf{Method} & PSNR & SSIM & $\Delta$E \\ \hline
  KNN  & 37.42 & 0.9882 & 3.43 \\ \hline
    MLP w/ X-Rite chart  & 38.91 & 0.9890 & 2.84 \\ \hline
        MLP w/ custom chart  & 38.93 & 0.9892 & 2.82 \\ \hline
    MLP w/ both  & \cellcolor{yellow!35}39.01 & \cellcolor{yellow!35}0.9892 & \cellcolor{yellow!35}2.82 \\ \hline
        Oracle  & 41.22 & 0.9872 & 1.95 \\ \hline

  \end{tabular}%
  }
\end{table}

Recent neural ISP methods (e.g.,~\cite{twostageisp,deepflexisp}) divide the ISP into multiple blocks, instead of replacing the entire ISP with one single monolithic DNN. This modular strategy is in line with traditional signal processing ISPs, and offers better interpretability and performance. We adopt a similar approach and divide the ISP into a white-balance (WB) block and a color-rendering block. We use our illumination-mapping approach to generate training data for the WB block and our sensor-mapping framework to synthesize data for the rendering block.

To generate synthetic data to train the WB block, we assume only a small number of images from sensor B are available -- 6 scenes under 8 illuminants for training, and 3 scenes under 2 illuminants for validation. The synthetic methods only have access to a dataset that is 5 times smaller than the oracle. 
Using our illumination-mapping MLP,
we augment each illuminant 4 times such that the number of images is the same as for the oracle. We use the C4 illuminant estimation network~\cite{c4} architecture for our WB block.

\begin{figure}[t]
  \centering
  \includegraphics[width=1.0\linewidth]{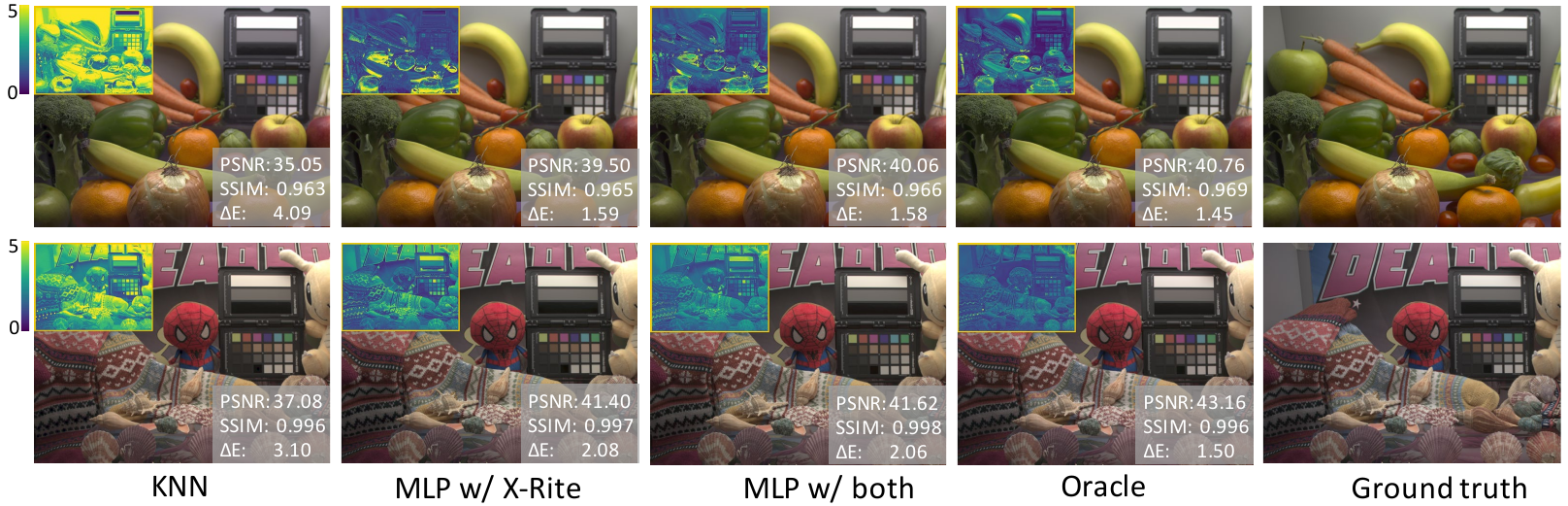}
  \caption{Qualitative results for the neural ISP task with the Canon DSLR as the target camera. Inset shows $\Delta$E error map.
    \label{fig:neuralisp}
  }
\end{figure}

For the rendering block, we map all images from sensor A to sensor B using our sensor mapping MLP.
We use a U-Net architecture for our rendering block. Implementation details can be found in the supplementary materials. For comparison, we also transform sensor A to B using the KNN 2NN w/ both approach reported in Table~\ref{tab:sensor-mapping-results}. To train the WB block of this KNN baseline, we also produce an augmented WB dataset using the KNN 1NN-2NN illumination-mapping method in Table~\ref{tab:illuminant-mapping-results}.  The rendering from RAW to sRGB is performed using the software ISP simulator in~\cite{sidd} for both real and synthetic RAW images. We show quantitative results in Table~\ref{tab:neural_isp} with the Samsung S22 Ultra smartphone as the source sensor and the Canon DSLR as the target sensor. The MLP-based method outperforms the KNN baseline by a sound margin. Qualitative results are shown in Fig.~\ref{fig:neuralisp} and Fig.~\ref{fig:Teaser}-B. 

\section{Discussion and conclusion}

This paper addresses an important yet under-investigated problem of RAW image manipulation for illumination and sensor augmentation. In particular, we focus on mapping functions that allow (1) a RAW image captured under a source illumination to appear as if it was captured under user-specified target illumination and (2) a RAW image captured by a source sensor for a given illumination to appear as if a target sensor under the same illumination captured it. 
To address this issue, we have captured a comprehensive RAW image dataset of 18 scenes with 390 illuminations for multiple cameras. Using this dataset, we showed that conventional approaches, such as simple channel scaling (i.e., diagonal matrix), nearest neighbor lookup, and U-Nets, are not sufficiently accurate for RAW image augmentation.   
We proposed an MLP-based method that predicts the necessary transforms for the illumination-to-illumination and sensor-to-sensor mapping problems.
We showed that this method significantly outperforms the baseline methods. We also demonstrated how our dataset and MLP-based mapping functions are helpful for low-level computer vision tasks targeting RAW image inputs such as neural rendering. 




\clearpage
\appendix

\title{Supplementary Material \\ Improved Mapping Between Illuminations and Sensors for RAW Images}
\author{} 

\vspace*{0.5cm}
This supplemental material provides additional details and results that were not included in the main paper.

\section{Dataset}
\label{sec:dataset_supp}

\begin{figure*}[b]
  \centering
  \includegraphics[width=\linewidth]{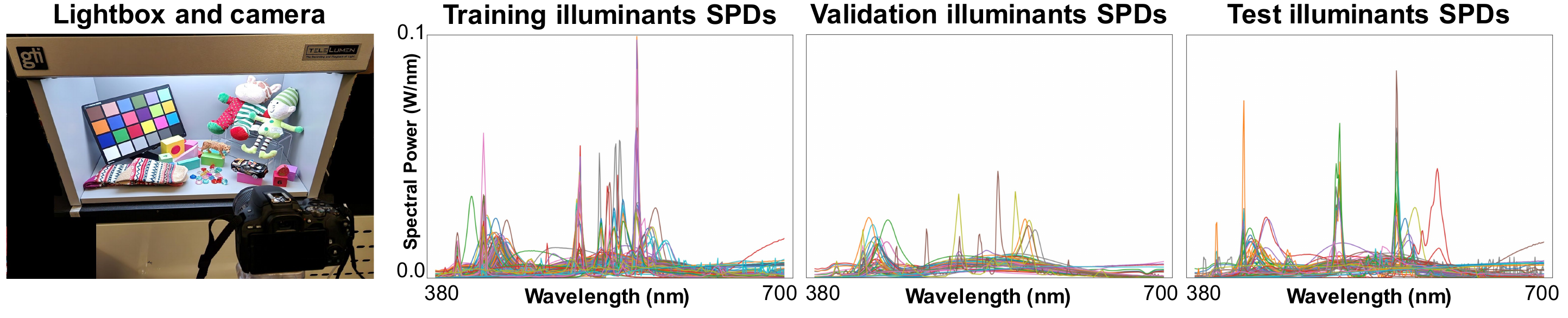}
  \caption{This figure shows our data capture setup. A GTI lightbox equipped with a Telelumen's tunable light source~\cite{telelumens} was used to produce spectral power distributions (SPDs) over the visible wavelengths (380-700 nm) sampled from illumination datasets representing natural and commercial lighting~\cite{SFU,PhotoLED,LSPDD,RealLightSource}. The SPDs for the illuminants used to illuminate our scenes for our training, validation, and testing images are shown.
    \label{fig:capture-setup}
  }
\end{figure*}

\begin{figure*}[!b]
\centering
\includegraphics[width=1.0\linewidth]{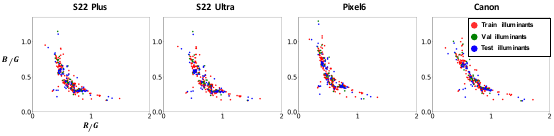}
\caption{The $[R/G,B/G]$ chromaticity plots of the illuminants across training, validation and test splits for the four cameras in our dataset.
\label{fig:illum_chromaticity}}
\end{figure*}

The first column of Fig.~\ref{fig:capture-setup} demonstrates our capture setup. The last three columns show the spectral power distributions of our three illuminant splits. Fig.~\ref{fig:illum_chromaticity} shows the [R/G,B/G] chromaticity plots of the illuminants in our dataset for the four cameras.


\begin{figure*}[!t]
  \centering
  \includegraphics[width=\linewidth]{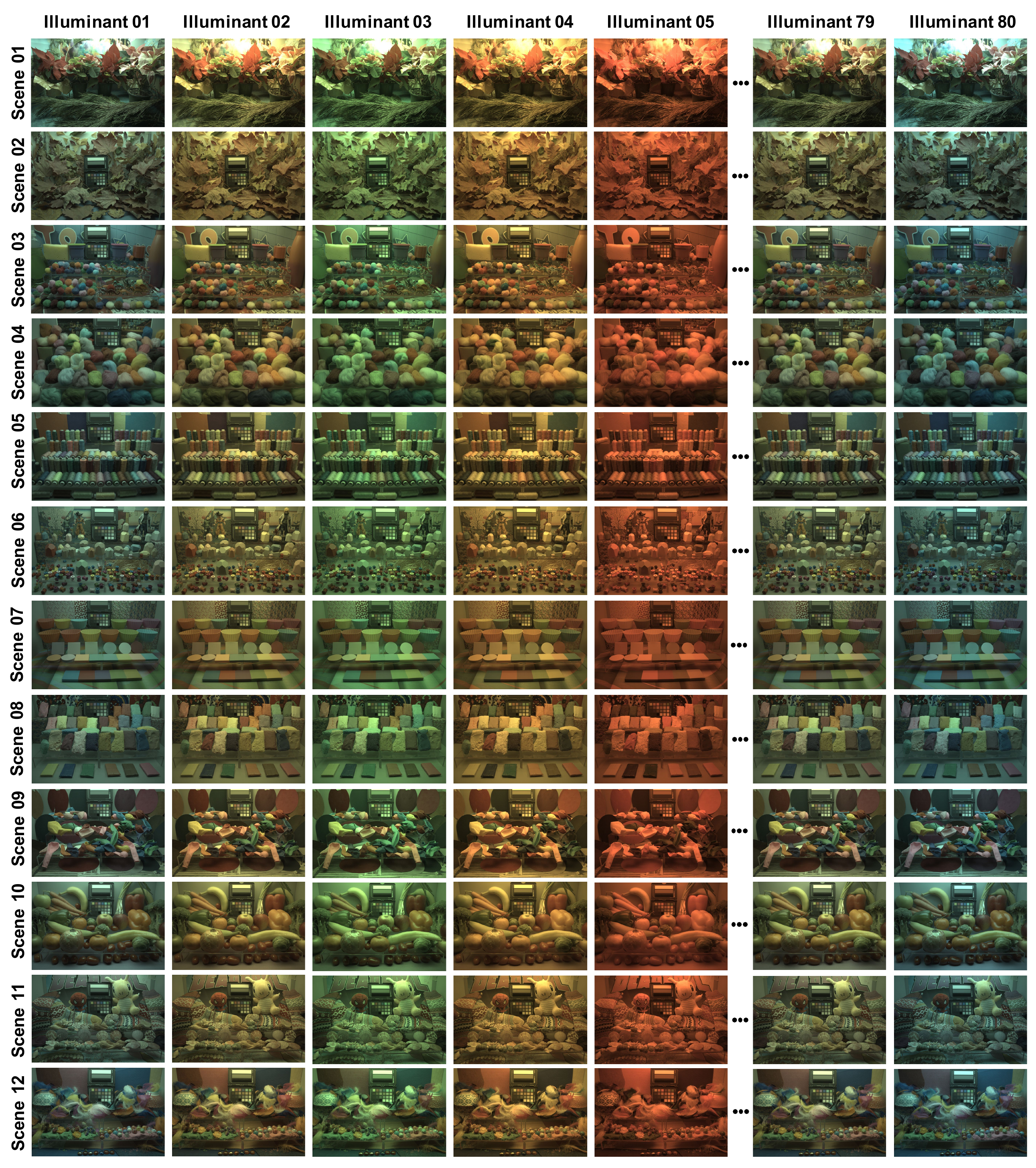}
  \caption{Illumination-to-illumination mapping dataset. We show representative examples from our 12 test scenes captured under 80 test illuminants using the Canon EOS Rebel SL2 DLSR.
    \label{fig:dataset_supp}
  }
\end{figure*}

In Fig. 2-A of our main paper, we provided representative examples of our illumination-to-illumination mapping dataset. Images captured using the Samsung S22 Plus smartphone camera were shown. As mentioned in the main paper, the same scenes under the same lighting conditions were also captured using the three other cameras in our dataset. In Fig.~\ref{fig:dataset_supp}, we provide additional examples of our dataset captured using the Canon EOS Rebel SL2 DSLR camera.

\section{Ablation studies}
We perform ablation studies on our MLP model and training data sampling scheme. All results reported in this section are for illuminant-to-illuminant mapping on the Samsung S22 Plus smartphone camera.

We conduct two ablation studies on our MLP architecture. We vary (1) the number of hidden layers (Table~\ref{tab:hidden-layers-ablation}) and (2) the number of neurons in each hidden layer (Table~\ref{tab:hidden-neurons-ablation}). Our choice of two hidden layers with 32 neurons in each layer produces the best performance.

As described in the main paper, we randomly sample 1000 pixels from the images at each training epoch to train our illumination mapping MLP. The ablation in Table~\ref{tab:random-pixels-ablation} shows this is an optimal choice. We hypothesize that random sampling is less prone to overfitting and yields better performance than the full image (last row of Table~\ref{tab:random-pixels-ablation}).

\section{Additional results}
\begin{figure*}[!t]
  \centering
  \includegraphics[width=\linewidth]{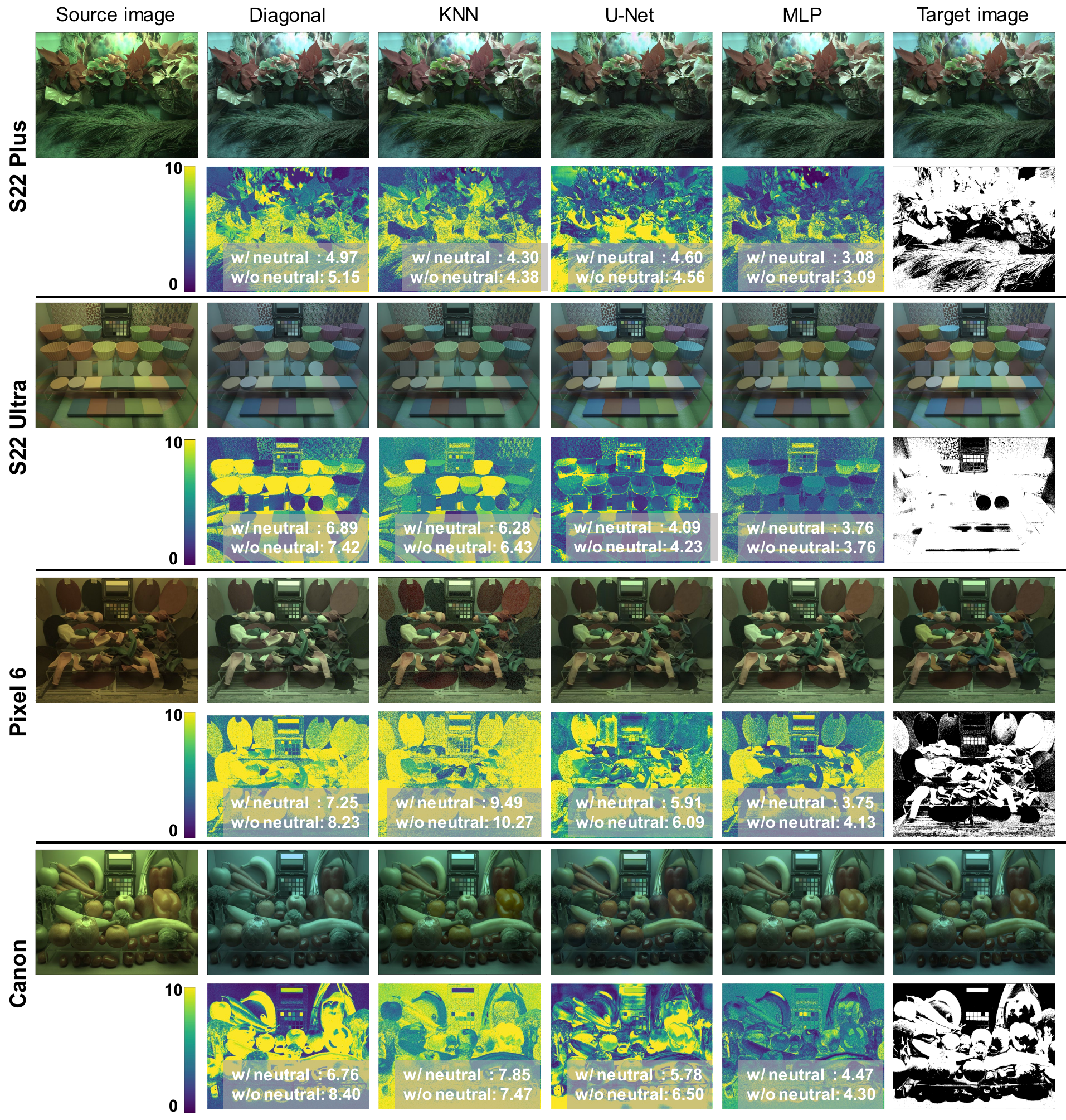}
  \caption{Qualitative comparisons between our method and competing approaches for illumination-to-illumination mapping. The error map shows per-pixel angular error, and insets provide the mean angular error. The binary mask shows the non-saturated, non-neutral pixels over which the angular error is computed in white color. The third KNN variant 2NN--1NN~\cite{lo2021clcc} is shown.
    \label{fig:illuminant-mapping-visual-supp}
  }
\end{figure*}

\begin{figure*}[!t]
  \centering
  \includegraphics[width=0.9\linewidth]{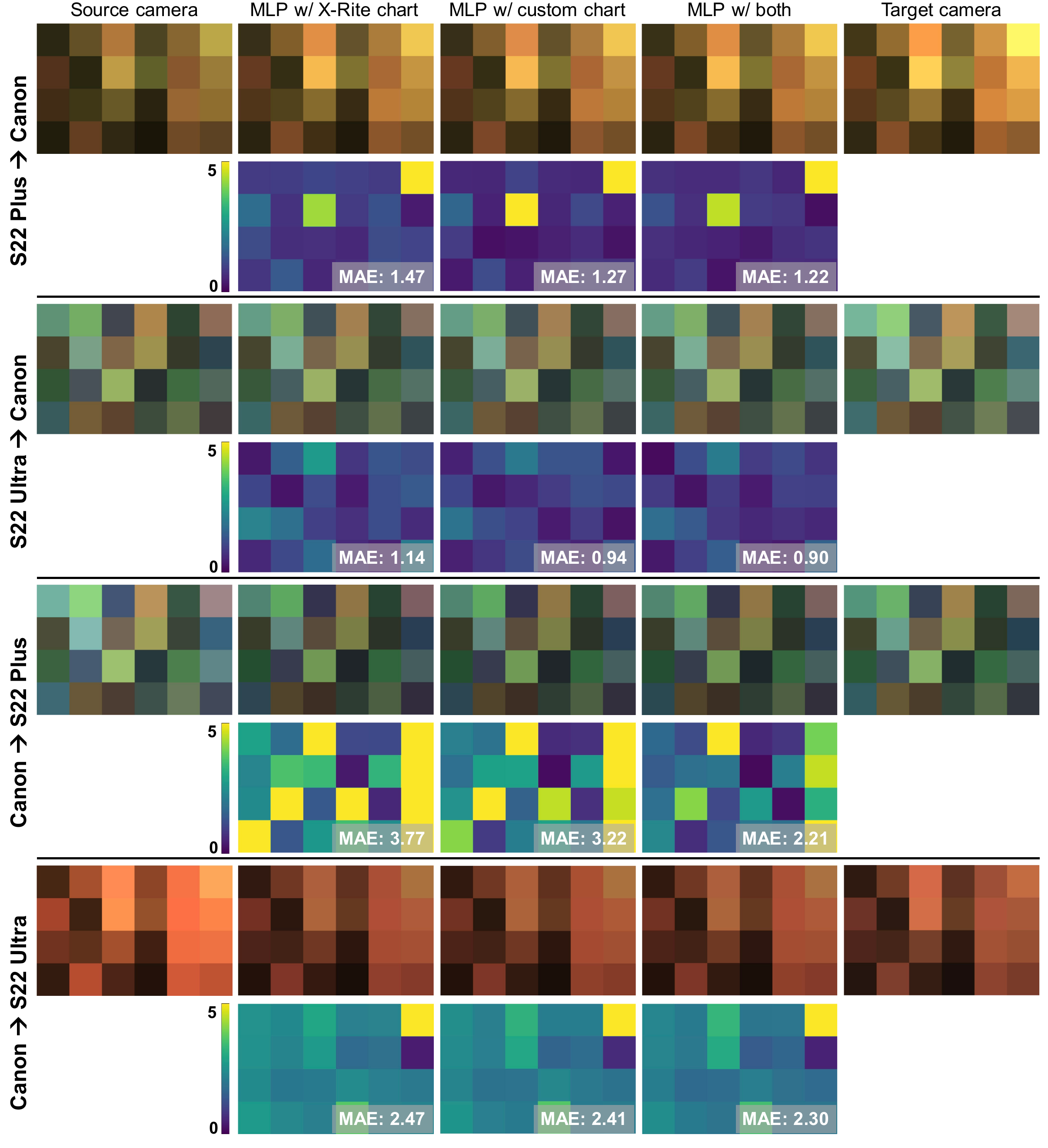}
  \caption{Qualitative comparisons between our MLP variants for sensor-to-sensor mapping. The error map shows per-pixel angular error, and insets provide the mean angular error.
    \label{fig:sensor-mapping-visual-supp}
  }
\end{figure*}

Fig.~\ref{fig:illuminant-mapping-visual-supp}  shows additional qualitative results, similar to Fig. 4-A of our main paper.  This figure compares our approach to competing methods for illumination-to-illumination mapping. 
Similarly, Fig.~\ref{fig:sensor-mapping-visual-supp} extends Fig. 4-B of our main paper for the sensor-to-sensor mapping task.

Similar to the neural ISP downstream task in Table 3 of our main paper, we perform another experiment with the Samsung S22 Plus smartphone as the source sensor and the Canon DSLR as the target sensor. Results are reported in Table~\ref{tab:neural_isp_supp}.

\begin{table}[t!]
\centering
  \caption{Ablation on the number of hidden layers in the MLP. Mean angular error (MAE) is reported.}
  \label{tab:hidden-layers-ablation}
  \resizebox{0.425\columnwidth}{!}{%
  \setlength{\tabcolsep}{6pt}
  \begin{tabular}{|l|c|c|}
  \hline
  \textbf{\# hidden layers} & \textbf{w/ neutral}          & \textbf{w/o neutral}         \\ \hline
  1  & 3.73                         & 3.81                         \\ \hline
  2 -- Ours & \cellcolor{yellow!35}3.63 & \cellcolor{yellow!35}3.67 \\ \hline
  3  & 3.90                         & 3.96                         \\ \hline
  \end{tabular}%
  }
\end{table}

\begin{table}[t!]
\centering
  \caption{Ablation on the number of neurons in each hidden layer in the MLP. MAE is reported.}
  \label{tab:hidden-neurons-ablation}
  \resizebox{0.375\columnwidth}{!}{%
  \setlength{\tabcolsep}{6pt}
  \begin{tabular}{|l|c|c|}
  \hline
  \textbf{\# neurons}   & \textbf{w/ neutral}          & \textbf{w/o neutral}         \\ \hline
  16  & 4.13                         & 4.15                         \\ \hline
  32 -- Ours & \cellcolor{yellow!35}3.63 & \cellcolor{yellow!35}3.67 \\ \hline
  64 & 3.68                         & 3.70                         \\ \hline
  \end{tabular}%
  }
\end{table}

\begin{table}[t!]
\centering
  \caption{Ablation on the number of random pixels chosen at each epoch of training. MAE is reported.}
  \label{tab:random-pixels-ablation}
  \resizebox{0.5\columnwidth}{!}{%
  \setlength{\tabcolsep}{6pt}
  \begin{tabular}{|l|c|c|}
  \hline
  \textbf{\# random pixels / epoch}          & \textbf{w/ neutral}          & \textbf{w/o neutral}         \\ \hline
  500  & 3.88 & 3.90 \\ \hline
  1000 -- Ours & \cellcolor{yellow!35}3.63 & \cellcolor{yellow!35}3.67 \\ \hline
  1500 & 4.00 & 4.07 \\ \hline
  All      & 4.09 & 4.14 \\ \hline
  \end{tabular}%
  }
  \end{table}

\begin{table}[t!]
\centering
  \caption{Quantitative results on the downstream task of neural rendering. Sensor A source: Samsung S22 Plus smartphone, sensor B target: Canon DSLR. A lower $\Delta$E~\cite{deltaE} is better.}
      \centering
  \label{tab:neural_isp_supp}
  \resizebox{0.375\columnwidth}{!}{%
  \setlength{\tabcolsep}{6pt}
  \begin{tabular}{|l|c|c|c|}
  \hline
  \textbf{Method} & PSNR & SSIM & $\Delta$E \\ \hline
  KNN  & 37.44 & 0.9881 & 3.42 \\ \hline
    MLP w/ both  & 38.82 & 0.9883 & 2.82 \\ \hline 
        Oracle  & 41.22 & 0.9872 & 1.95 \\ \hline

  \end{tabular}%
  }
\end{table}

\begin{figure*}[!t]
  \centering
  \includegraphics[width=1\linewidth]{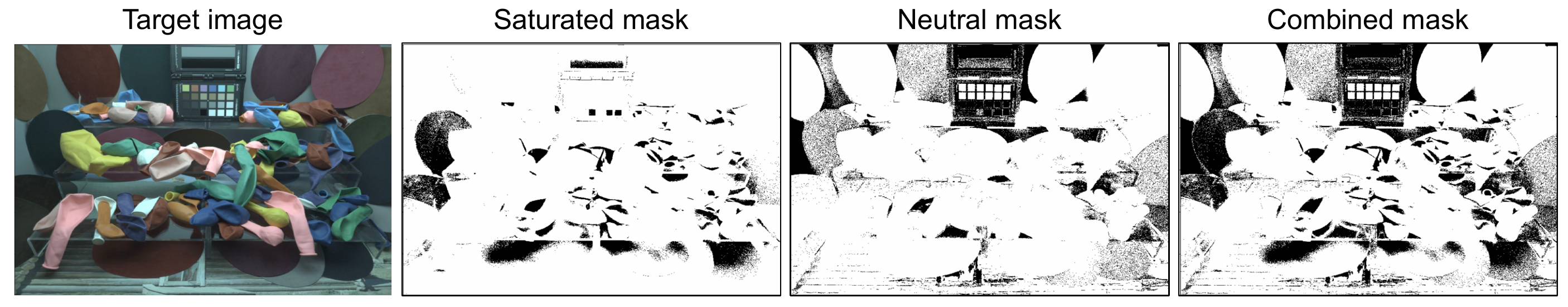}
  \caption{This figure shows a target image from one test scene captured using the S22 Ultra smartphone and its three corresponding masks. Saturated mask: non-saturated pixels are white. Neutral mask: non-neutral pixels are white.
    \label{fig:saturated-neutral-masks}
  }
\end{figure*}

\begin{figure*}[!t]
  \centering
  \includegraphics[width=1\linewidth]{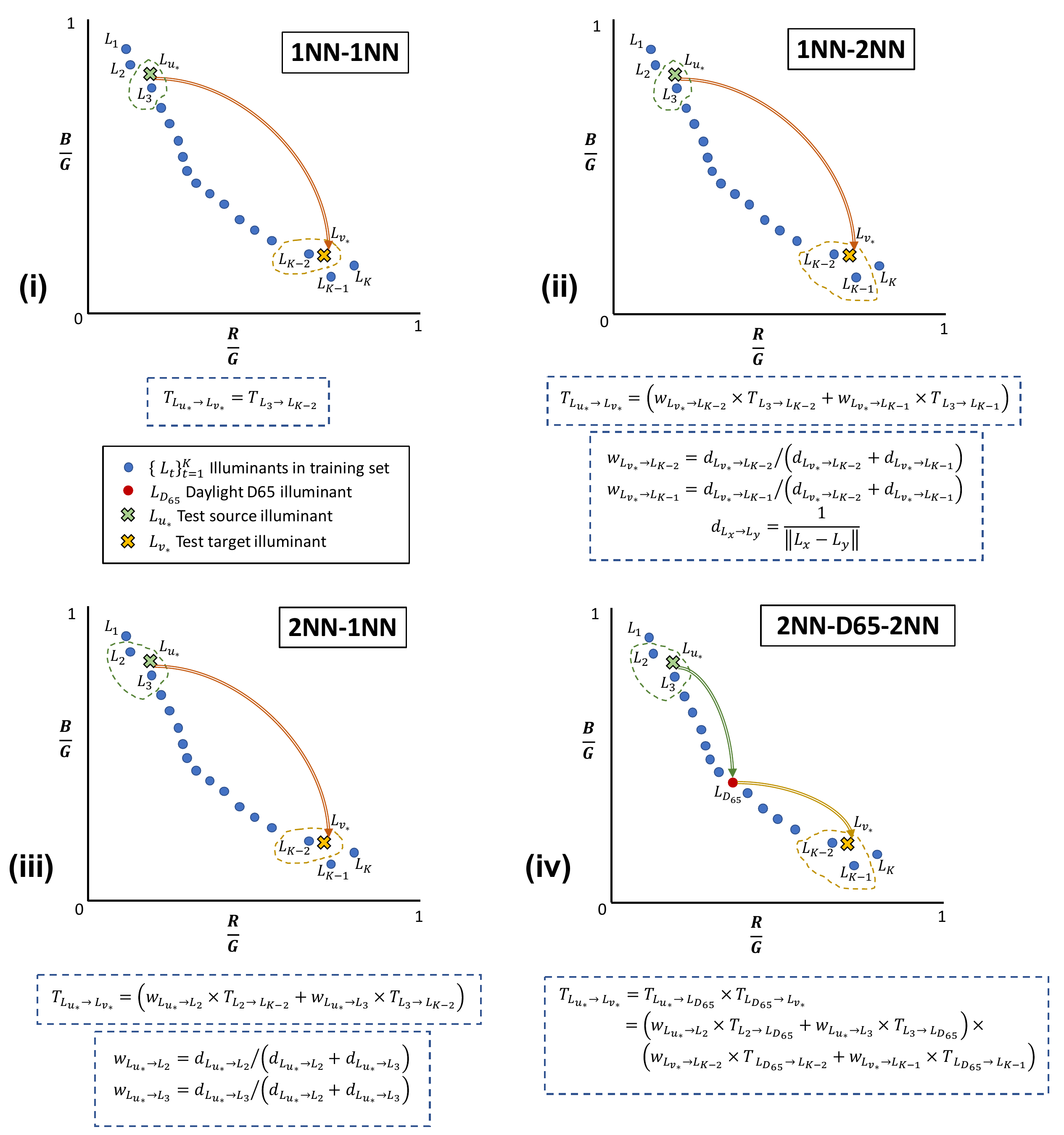}
  \caption{An illustration of the four KNN variants used as comparison baselines for our illumination-to-illumination mapping task. The four variants, depending on the value of K used to interpolate the source and target illuminants, are (i) 1NN--1NN, (ii) 1NN--2NN, (iii) 2NN--1NN, and (iv) 2NN--D65--2NN. The sensor's RAW color space is shown in the $\left[ \frac{R}{G}, \frac{B}{G} \right]$ chromaticity plots. The training illuminants are the set $\{ L_t \}_{t=1}^K$ which includes the daylight D65 illuminant. The transformation matrices between all illuminant pairs in the training set are assumed to be precomputed and available at test time. The source and target illuminants at test time are denoted as $L_{u_{*}}$ and $L_{v_{*}}$, respectively. The matrix $T_{L_{u_{*}} \rightarrow L_{v_{*}}}$ is the desired transformation that maps the image under the source illumination $L_{u_{*}}$ to the target illumination $L_{v_{*}}$.
    \label{fig:knn}
  }
\end{figure*}

\section{Implementation details}
\subsection{Illumination and sensor mapping}
As mentioned in the experiments section of the main paper, we excluded saturated pixels from our MAE computation since angular errors are not meaningful if one or more color channels are saturated. We also introduced a split of with and without neutral colors to study the accuracy of full-color correction.  Fig.~\ref{fig:saturated-neutral-masks} demonstrates an example target image and its corresponding saturated, neutral, and combined masks.

For the U-Net~\cite{unet} baseline experiments for illumination mapping, we trained all U-Net models for 50 epochs with a learning rate of 0.001 and a batch size of 128.
We used the Adam optimizer~\cite{adam} with $\beta_1=0.9$ and $\beta_2=0.999$. The U-Net is trained to predict a residual layer added to its input to obtain the output image. Our experiments found that residual prediction produces better results than directly predicting the output image.

For fairness of comparison with our MLP, we trained and validated the U-Net on a single scene.
Since only one scene is available, we trained on patches of size $64 \times 64$ pixels to avoid dependence on the scene's spatial structure. 
Our MLP uses only a fraction of the parameters compared to the U-Net, and is trained to predict the parameters of the transformation, as against the image itself. Thus, the single scene provides sufficient training data for our MLP. However, in our experiments, we found that training a U-Net on a single scene, even with thousands of illuminant pairs, does not always yield a robust, generalizable model.

In the experiments section of our main paper, we described four K-nearest neighbor (KNN) variants used as baselines for our experiments on illumination-to-illumination mapping. Fig.~\ref{fig:knn} illustrates how each variant's transformation matrix is estimated. The KNN comparison for sensor-to-sensor mapping in Table 2 of our main paper, although not included in the figure, follows a similar implementation.

\subsection{Neural ISP}
As mentioned in the main paper, for our neural ISP experiment, we used the C4 illuminant estimation network~\cite{c4} as the architecture for our WB block and a U-Net~\cite{unet} architecture for our rendering block. We used the official implementation from the authors with their recommended hyper-parameters to train the C4 network.
For the U-Net, we trained for 4000 epochs with a learning rate of 0.0001 and a batch size of 24 on patches of size $256\!\times\!256$ pixels.
We used the Adam optimizer~\cite{adam} with $\beta_1=0.9$ and $\beta_2=0.999$. Note that we do not include a denoiser in our neural ISP because our focus is on color reproduction accuracy and all images in our dataset were captured at the lowest ISO setting. Demosaicing was also applied as a pre-processing step, as mentioned in the experiments section of our main paper. For the neural ISP experiment, we also reported the $\Delta$E value~\cite{deltaE} in addition to PSNR and SSIM since $\Delta$E is widely used to measure changes in visual perception between two colors in sRGB space.

Finally, all our experiments are repeated four times and the average is reported in our tables.

\section{Choice of evaluation metric}
To quantitatively examine the accuracy of our mapping functions, we use angular error as the evaluation metric. Our focus in both tasks---illumination-to-illumination mapping and sensor-to-sensor mapping---is on color reproduction accuracy in the RAW space, which can be accurately measured using angular error. From the perspective of color reproduction, the magnitude of the RAW R, G, B response does not matter; any uniform scale factor (i.e., $\alpha$R, $\alpha$G, $\alpha$B) of the camera's response represents the same illumination color. The three-dimensional RAW R, G, B vector can be interpreted as a color \textit{ray}. The change in $\alpha$ is due to varying radiant power of the scene's illumination and camera settings, such as exposure and ISO gain.

\begin{figure*}[!t]
  \centering
  \includegraphics[width=1\linewidth]{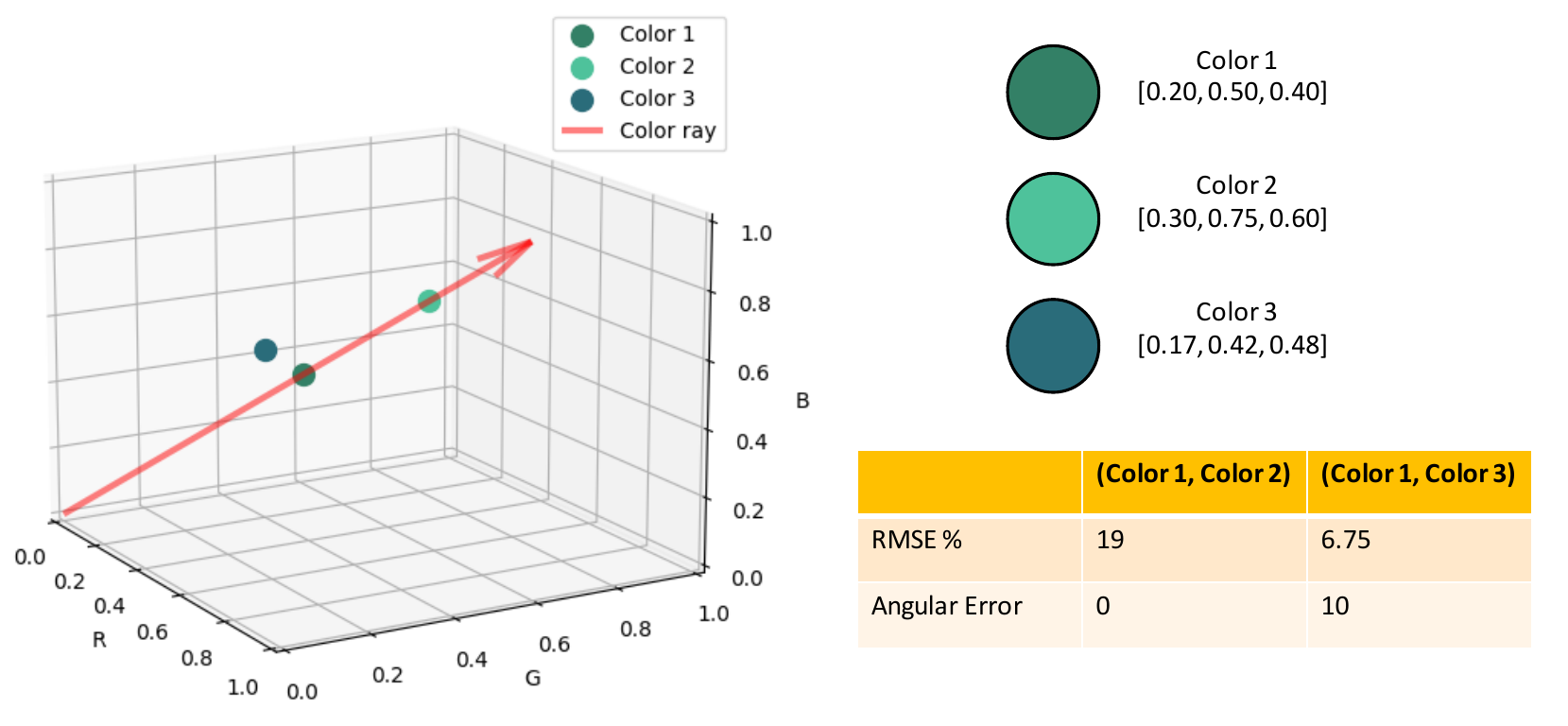}
  \caption{A comparison between RMSE and angular error as an evaluation metric for color reproduction accuracy. The angular error is zero between two colors (Color 1 and Color 2) that lie on the same color ray (i.e., two colors that are scaled versions of each other), but the RMSE, which depends on the magnitude of the color vectors, is high. Moreover, RMSE may fail to distinguish between two different colors (Color 1 and Color 3) if the vectors have similar magnitudes. This makes RMSE unsuitable as an evaluation metric for color recovery.
    \label{fig:rmse}
  }
\end{figure*}

Even though we carefully control the illumination in the scene during data capture, minor fluctuations in the camera's auto-exposure algorithm can still result in global changes in brightness between frames. This can lead to a global but unknown scale factor $\alpha$ between the estimated image and the ground truth. The angular error will correctly ignore such brightness differences between pixel intensities, but metrics such as root mean squared error (RMSE) may report a large error value. This is illustrated in Fig.~\ref{fig:rmse} using three RAW colors. Color 1 and Color 2 lie along the same color ray---Color 2 is 1.5 times brighter than Color 1. The angular error between these two colors is zero, but the RMSE is approximately 20\%. Color 3 represents a different color that does not lie on the same ray as Color 1. However, the RMSE between Color 1 and Color 3 is roughly three times smaller than the RMSE between Color 1 and Color 2. On the other hand, the angular error metric reports a large value of 10 degrees, rightly signifying that Colors 1 and 3 are two distinct colors that are far apart.

\bibliography{egbib}

\begin{thebibliography}{10}
\newcommand{\enquote}[1]{``#1''}

\bibitem{deltaE}
G.~Sharma and R.~Bala, \emph{Digital color imaging handbook} (CRC Press, 2013), 2nd ed.

\bibitem{chatfield2014return}
K.~Chatfield, K.~Simonyan, A.~Vedaldi, and A.~Zisserman, \enquote{Return of the devil in the details: Delving deep into convolutional nets,} {\protect\JournalTitle{BMVC}}  (2014).

\bibitem{cubuk2018autoaugment}
E.~D. Cubuk, B.~Zoph, D.~Mane, \emph{et~al.}, \enquote{Autoaugment: Learning augmentation policies from data,} {\protect\JournalTitle{CVPR}}  (2019).

\bibitem{doersch2015unsupervised}
C.~Doersch, A.~Gupta, and A.~A. Efros, \enquote{Unsupervised visual representation learning by context prediction,} in \emph{ICCV,}  (2015).

\bibitem{kalantari2017deep}
N.~K. Kalantari and R.~Ramamoorthi, \enquote{Deep high dynamic range imaging of dynamic scenes,} {\protect\JournalTitle{ACM ToG}} \textbf{36}, 1--12 (2017).

\bibitem{krizhevsky2017imagenet}
A.~Krizhevsky, I.~Sutskever, and G.~E. Hinton, \enquote{Imagenet classification with deep convolutional neural networks,} {\protect\JournalTitle{Communications of the ACM}} \textbf{60}, 84--90 (2017).

\bibitem{lee2017unsupervised}
H.-Y. Lee, J.-B. Huang, M.~Singh, and M.-H. Yang, \enquote{Unsupervised representation learning by sorting sequences,} in \emph{ICCV,}  (2017).

\bibitem{movshovitz2016useful}
Y.~Movshovitz-Attias, T.~Kanade, and Y.~Sheikh, \enquote{How useful is photo-realistic rendering for visual learning?} in \emph{ECCV,}  (2016).

\bibitem{redmon2016you}
J.~Redmon, S.~Divvala, R.~Girshick, and A.~Farhadi, \enquote{You only look once: Unified, real-time object detection,} in \emph{CVPR,}  (2016).

\bibitem{jung2019imgaug}
{Imgaug Library}, \url{https://imgaug.readthedocs.io/en/latest/} (2024). Accessed: 2025-02-26.

\bibitem{hu2017fc}
Y.~Hu, B.~Wang, and S.~Lin, \enquote{{FC4}: Fully convolutional color constancy with confidence-weighted pooling,} in \emph{CVPR,}  (2017).

\bibitem{lo2021clcc}
Y.-C. Lo, C.-C. Chang, H.-C. Chiu, \emph{et~al.}, \enquote{{CLCC}: Contrastive learning for color constancy,} in \emph{CVPR,}  (2021).

\bibitem{abdelhamed2021leveraging}
A.~Abdelhamed, A.~Punnappurath, and M.~S. Brown, \enquote{Leveraging the availability of two cameras for illuminant estimation,} in \emph{CVPR,}  (2021).

\bibitem{nguyen2015raw}
N.~H.~M. Rang, D.~K. Prasad, and M.~S. Brown, \enquote{Raw-to-{R}aw: {M}apping between image sensor color responses,} in \emph{CVPR,}  (2014).

\bibitem{gti}
{GTI Graphics Technology Inc.}, \url{https://www.gtilite.com/products/color-matching-systems/gti-minimatcher-series/} (2024). Accessed: 2025-02-26.

\bibitem{telelumens}
{Telelumen}, \url{https://www.telelumen.com/products/} (2024). Accessed: 2025-02-26.

\bibitem{SFU}
K.~Barnard, L.~Martin, B.~Funt, and A.~Coath, \enquote{A data set for color research,} {\protect\JournalTitle{Color Research and Application}} \textbf{27}, 147--151 (2002).

\bibitem{PhotoLED}
A.~{Kokka}, T.~{Poikonen}, P.~{Blattner}, \emph{et~al.}, \enquote{Development of white {LED} illuminants for colorimetry and recommendation of white {LED} reference spectrum for photometry,} {\protect\JournalTitle{Metrologia}} \textbf{55}, 526 (2018).

\bibitem{LSPDD}
J.~Roby, M.~Aub{\'e}, A.~Morin-Paulhus, \emph{et~al.}, \enquote{{LSPDD}: Lamp spectral power distribution database,} \url{https://lspdd.org/database/} (2024). Accessed: 2025-02-26.

\bibitem{RealLightSource}
M.~Royer, \enquote{Real light source {SPD}s and color data for use in research,}  (2024). {P}acific Northwest National Laboratory.

\bibitem{cheng2015}
D.~Cheng, B.~Price, S.~Cohen, and M.~S. Brown, \enquote{Beyond white: Ground truth colors for color constancy correction,} in \emph{ICCV,}  (2015).

\bibitem{adam}
D.~P. Kingma and J.~Ba, \enquote{Adam: A method for stochastic optimization,} in \emph{ICLR,}  (2015).

\bibitem{unet}
O.~Ronneberger, P.~Fischer, and T.~Brox, \enquote{U-{N}et: Convolutional networks for biomedical image segmentation,} in \emph{MICCAI,}  (2015).

\bibitem{twostageisp}
Z.~Liang, J.~Cai, Z.~Cao, and L.~Zhang, \enquote{Camera{N}et: {A} two-stage framework for effective camera {ISP} learning,} {\protect\JournalTitle{IEEE Transactions on Image Processing}} \textbf{30}, 2248--2262 (2021).

\bibitem{deepflexisp}
S.~Liu, C.~Feng, X.~Wang, \emph{et~al.}, \enquote{Deep-{F}lex{ISP}: A three-stage framework for night photography rendering,} in \emph{CVPR Workshop,}  (2022).

\bibitem{c4}
H.~Yu, K.~Chen, K.~Wang, \emph{et~al.}, \enquote{Cascading convolutional color constancy,} in \emph{AAAI,}  (2020).

\bibitem{sidd}
A.~Abdelhamed, S.~Lin, and M.~S. Brown, \enquote{A high-quality denoising dataset for smartphone cameras,} in \emph{CVPR,}  (2018).

\end{thebibliography}

\end{document}